\documentclass[10pt,twocolumn,letterpaper]{article}

\usepackage{cvpr}
\usepackage{times}
\usepackage{epsfig}
\usepackage{graphicx}
\usepackage{amsmath}
\usepackage{amssymb}
\usepackage{bbm}

\usepackage[pagebackref=false,breaklinks=true,colorlinks,bookmarks=false]{hyperref}

\cvprfinalcopy 

\def\cvprPaperID{****} 

\begin{document}

\title{Training a Fully Convolutional Neural Network to Route Integrated Circuits}

\newcommand*\samethanks[1][\value{footnote}]{\footnotemark[#1]}

\author{
  Sambhav R. Jain~\thanks{Corresponding author: sambhav@alumni.stanford.edu}~~\thanks{Indicates equal contribution.}\\
  Oracle America\\
  \and
  Kye Okabe~\samethanks \\
  Stanford University \\
}

\maketitle

\vspace{-0.05in}
\begin{abstract}
\vspace{-0.05in}
\noindent We present a deep, fully convolutional neural network that learns to route a circuit layout `net' with appropriate choice of metal tracks and wire class combinations. Inputs to the network are the encoded layouts containing spatial location of pins to be routed. After 15 fully convolutional stages followed by a score comparator, the network outputs 8 layout layers (corresponding to 4 route layers, 3 via layers and an identity-mapped pin layer) which are then decoded to obtain the routed layouts. We formulate this as a binary segmentation problem on a per-pixel per-layer basis, where the network is trained to correctly classify pixels in each layout layer to be `on' or `off'. To demonstrate learnability of layout design rules, we train the network on a dataset of 50,000 train and 10,000 validation samples that we generate based on certain pre-defined layout constraints. Precision, recall and $F_1$ score metrics are used to track the training progress. Our network achieves $F_1\approx97\%$ on the train set and $F_1\approx92\%$ on the validation set. We use PyTorch for implementing our model. Code is made publicly available\footnote{Code: https://github.com/sjain-stanford/deep-route}.
\end{abstract}

\vspace{-0.15in}
\section{Introduction}
\vspace{-0.05in}

\begin{figure}[t]
  \centering
  \includegraphics[width=\linewidth]{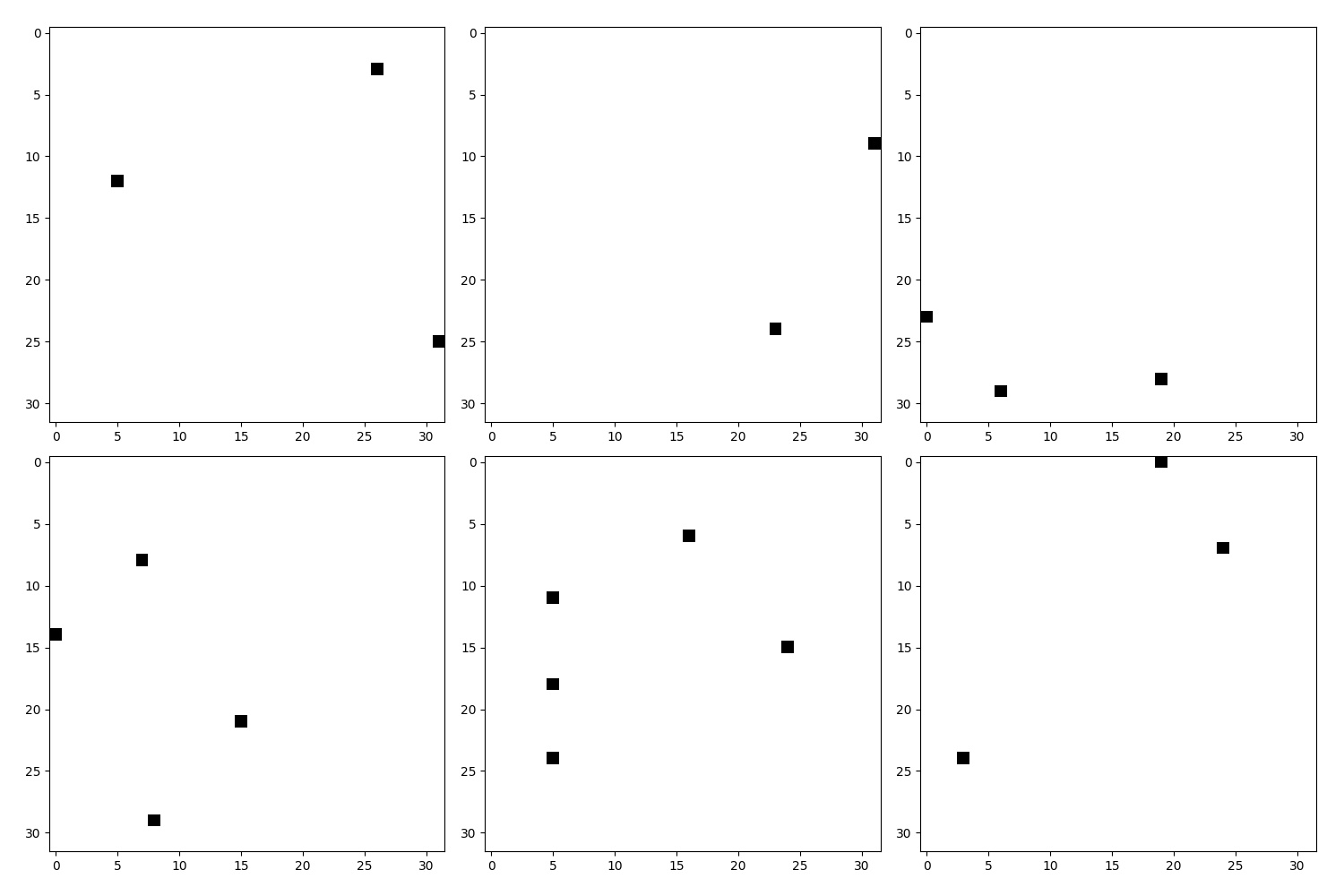}\\*[1mm]
  \includegraphics[width=\linewidth]{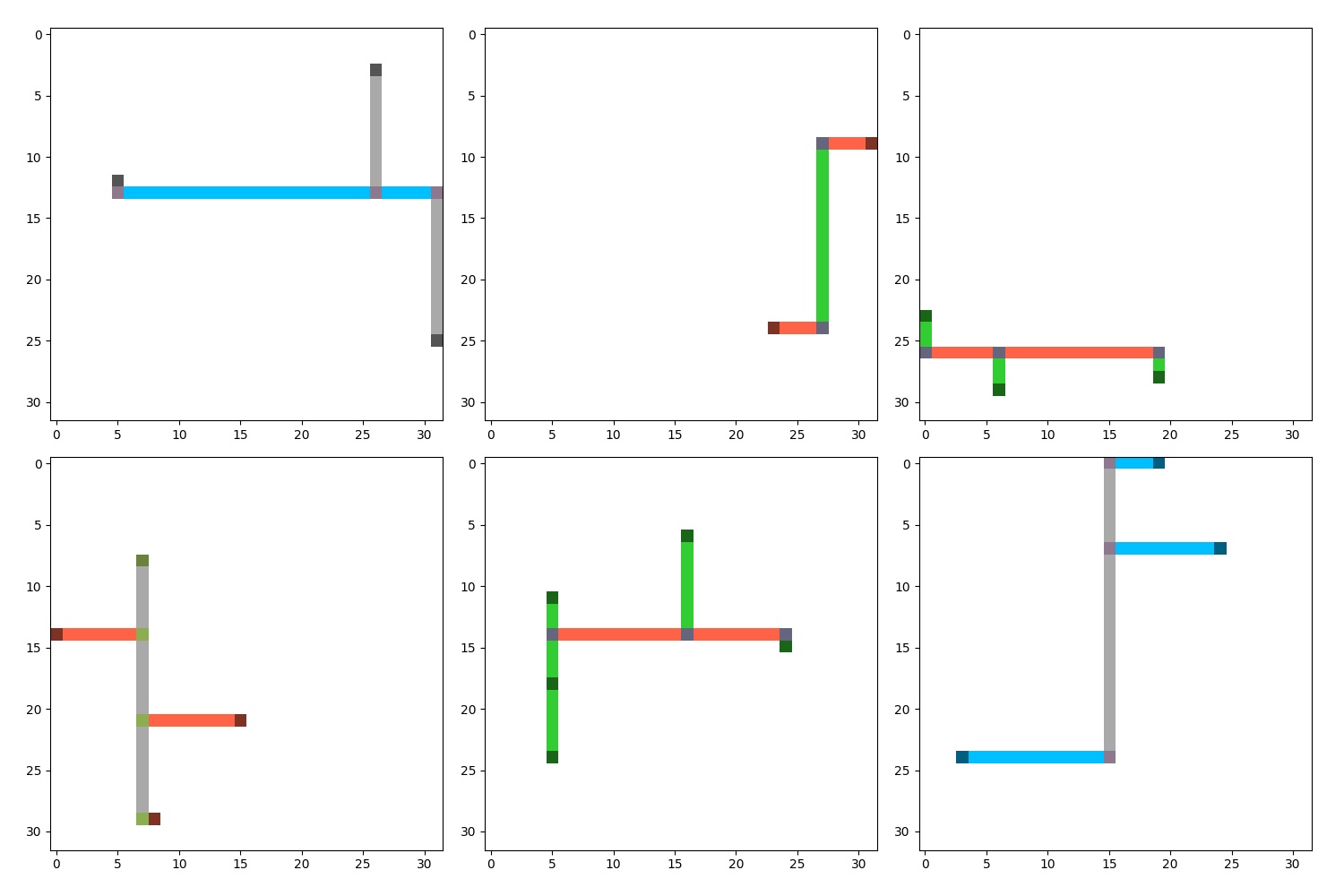}
  \caption{Training samples from our dataset. Top row shows input data containing pins for a single net case. Bottom row shows input labels (decoded to 8-bit RGB) containing pins, routes, and vias. Pins are routed using two wire classes (branch-leg). Wire class color coding: $M_3$=green, $M_4$=red, $M_5$=grey, $M_6$=blue.}
  \label{fig:sample_data}
  \vspace{-0.25in}
\end{figure}

\noindent Routing is a complex \textit{spatial} optimization problem in the physical design of integrated circuits (ICs) that is known to be NP-complete in most cases \cite{sherwani1995routing,pal2000multi,kramer82,Szymanski85}. The task is to optimally connect circuit segments spanning multiple layout hierarchies and multiple wire classes, while complying to a strict set of design rules dictated by the foundry's process design kit (PDK). The quality of routing determines circuit performance (frequency), reliability, and can also impact area. Depending on the type of circuit (ASIC / memory / processor), the routing objective may prioritize one over the other, but in general the expectation is to:
\begin{itemize}
  \itemsep-0.4em 
  \item Minimize path delay (resistance/capacitance)
  \item Minimize congestion (wire crowding, hot spots)
  \item Maximize routability (availability of tracks)
  \item Maximize repeatability
\end{itemize}

Formally, a circuit layout consists of several sub-blocks (or \textit{cells}) with input/output ports (or \textit{pins}). Connections between pins use channels (or \textit{tracks}) in several metal layers that run orthogonal to one another, separated by insulating layers with connecting \textit{vias}. If we treat cells as edges, then a \textit{node} in a circuit graph corresponds to a \textit{net} in the layout, which connects the driver's output pin with the input pin(s) of the receiver(s). Traditionally, when the circuit sizes were tractable, routing was primarily a manual task. However, in the past couple decades with exploding circuit sizes and more than a few billion transistors to be routed, the use of auto-routers is indispensable. Shrinking technology nodes and more stringent design rules further impact routability. Commercial electronic design automation (EDA) tools try to tackle this non-linear optimization problem using various algorithmic approaches \cite{sherwani1995routing} such as (i) exponential algorithms to exhaust the search space for a solution, (ii) heuristic algorithms, (iii) optimal algorithms for special cases of the problem, or (iv) approximation techniques. These approaches are iterative in nature and rely on a continually evolving/changing set of design rules. The routed layouts may further require significant manual effort and fine tuning to improve the non-optimal routes.

We explore a learning-based approach wherein we train a deep, fully convolutional network (FCN) to route a layout net while relying on its ability to learn implicit design rules from the training data. In order to demonstrate learnability of layout design rules by the network, we pre-define a set of basic constraints (not specific to any technology node) that are embedded in the ground truth layouts generated for a single net case. The network is trained on our dataset of 50,000 train and 10,000 validation samples. Input data (containing pins) and labels (containing pins and routes) are first encoded on a binary basis per-pixel per-layer (detailed in Section~\ref{sec:dataset}), before feeding to the 15 stage FCN. FCN outputs go through a score comparator to give 8-layer encoded layouts (4 route layers, 3 via layers, and 1 pin layer). We decode this to 8-bit RGB for visualization. The various architectural choices of the FCN are covered in depth in Sections~\ref{sec:model} and \ref{sec:experiments}. The network shows ability to learn (i) the identity-map for pins, (ii) optimal track locations for routes, (iii) suitable wire class combination, and (iv) via locations.

\vspace{-0.05in}
\section{Related Work}
\vspace{-0.05in}

\noindent The task of producing routes using orthogonal layers of parallel channels and vias at intersections is often tackled heuristically, since an optimal solution does not exist for this NP-complete problem \cite{sherwani1995routing,pal2000multi,kramer82,Szymanski85}. Previous work in this field \cite{Yoshimura82,Rivest82,Burstein83,Yoshimura84,Wang90,Zajc91} mostly rely on explicit rule-based algorithms to tackle parts of this complex task. For instance, Zajc \etal \cite{Zajc91} proposed using hierarchical algorithms for automatic routing of interconnects in two layer channels. Rivest \etal \cite{Rivest82} show a "greedy" heuristic channel router assuming all pins and wiring sit on a common grid. The other class of routers, which have received attention lately, are objective function based. Alpert \etal \cite{Alpert93} proposed a combination of minimum spanning tree (Prim) and shortest path tree (Dijkstra) objectives for performance driven global routing. Constructing routing tree with different objectives was also explored, such as timing objective \cite{Song14}, buffer insertion \cite{Lillis96} and wire sizing objective \cite{Tang01}, congestion and thermal objective \cite{Roy14,Ivanov16}. A completely different objective for diagonal routing as opposed to the orthogonal (Manhattan) routing was also proposed \cite{Samanta06}.

One of the very first ideas of using neural networks for circuit routing came from Green and Noakes \cite{Green89}, who divided the routing task into several stages and combined multiple small back propagation networks to form a complex neural system. They proposed to restrict the routing task to predefined areas and slide the context window to complete routes in sections. This divide and conquer approach helped reduce the complexity of their artificial neural network router. Our work takes inspiration from the recent advances in convolutional neural networks (CNN) \cite{Fukushima80,Lecun95,Krizhevsky12} which are better suited for visual tasks as they preserve spatial information in the inputs. Similar to \cite{Green89}, we fix the routing window to a predefined layout size, however, in contrast to \cite{Green89}, we develop a single, end-to-end, deep network using convolutions, which holistically learns multiple design rules during training, and is able to route using different wire class combinations, depending on the spatial spread of the pins. To our best knowledge, this is the first attempt at routing a circuit layout net using a convolutional neural network.

\vspace{-0.05in}
\section{Dataset} \label{sec:dataset}
\vspace{-0.05in}

\noindent \textbf{Overview.} We develop our own dataset owing to the lack of a publicly available layout dataset, and the need to use simplistic design rules to assess learnability by the network (feasibility). In this section we discuss the design choices and constraints used when generating 50,000 layouts for training and 10,000 layouts for validation. Each layout sample contains both data (pins only) and labels (pins, routes, vias) for a single net (see Figure~\ref{fig:sample_data}). The image is pixel-wise binary encoded in 8 layers of the layout, viz. [$pin$, $M_{3}$, $Via_{3}$, $M_{4}$, $Via_{4}$, $M_{5}$, $Via_{5}$, $M_{6}$]. So each pixel in a given layer is either '1' or '0', indicating the presence or absence of the layer at that spatial location. Our reasoning behind this encoding scheme is discussed in Section~\ref{sec:colorencoding}. Thus training data are stored as tensors of shape $N \times 1 \times H \times W$, and labels as tensors of shape $N \times 8 \times H \times W$, where $H=W=32$. The layouts were sized $32\times32$ pixels, which we found computationally feasible. We use chunked HDF5 \cite{hdf5} when storing and loading this large dataset to avoid memory bottlenecks.

\vspace{+0.05in}
\noindent \textbf{Design rules.} These design rules are fundamental to traditional layout design and do not correspond to any specific technology node.
\vspace{-0.05in}
\begin{enumerate}
  \itemsep-0.4em 
  \item $M_{2n}$ tracks run horizontally.
  \item $M_{2n+1}$ tracks run vertically.
  \item $M_{n}$ contacts $M_{n+1}$ through $Via_{n}$.
\end{enumerate}

\noindent \textbf{Design choices.} We made the following design choices to reduce the complexity of network implementation. These could be scaled as needed given sufficient resources.
\vspace{-0.05in}
\begin{enumerate}
  \itemsep-0.4em 
  \item Maximum $n_{pins} = 5$.
  \item Layout context window = $11\mu m \times 11 \mu m$.
  \item Total allowed route layers = $4$; total via layers = $3$.
  \item Routes are limited to two wire classes (branch-leg) (e.g. $[M_{3},M_{4}]$ or [$M_{4},M_{5}]$ or $[M_{5},M_{6}]$).
  \item Metal tracks use pixel grid (one track per pixel).
  \item Pin layer is lower than all route layers, and routes need not drop vias down to the pin layer.
\end{enumerate}
\vspace{-0.05in}
\noindent Higher metals normally have less wire resistance per unit length compared to lower metals, due to their large cross sectional area and/or better material properties. However, higher metals also require additional vias to jump up/down to/from higher layers, which adds to the total resistance. As a result, there is a break-even route distance above which a higher metal is preferred. Moreover, for a given layout size, depending on wire performance data there are only a limited set of wire classes which make sense for routing. Using higher or lower wire classes than necessary would be non-optimal. In our case, we use empirical wire and via resistance data\footnote{Not released due to proprietary reasons} for four such wire classes (say $M_{3}$ to $M_{6}$). We select $11\mu m$ as the size of the context window as it gave us a balanced dataset among the three wire class combinations, viz. $[M_{3},M_{4}]$, $[M_{4},M_{5}]$, $[M_{5},M_{6}]$. By defining a context window for routing, we let the possibility open to have it slide over the entire layout to complete routing in segments, similar to \cite{Green89}, but we leave the slide-and-route implementation to future work.

\vspace{+0.05in}
\noindent \textbf{Route algorithm.} For each layout we first sample (random uniform) $n_{pins}$ from $\{2,3,4,5\}$ and $(x, y)$ co-ordinates for each pin from $\{0-31\}$ (since pixel grid is $32\times32$). For a given pin configuration, the direction with the largest spread of pins is chosen as dominant. The dominant direction uses a branch, while the non-dominant direction uses legs to connect individual pins to the branch. The choice of metals for branch and legs is done so as to optimize the combined wire and via resistance, as explained earlier. From Figure~\ref{fig:sample_data}, we can qualitatively see that shorter routes use lower metals while longer routes use higher metals. We can also see that a branch is always assigned to the dominant direction, and legs to the other direction. For visualization, the routed layouts are decoded to 8-bit RGB as follows: $M_{3}$ as green, $M_{4}$ as red, $M_{5}$ as grey and $M_{6}$ as blue.



\begin{figure*}[t]
  \vspace{-0.15in}
  \centering
  \includegraphics[width=\linewidth]{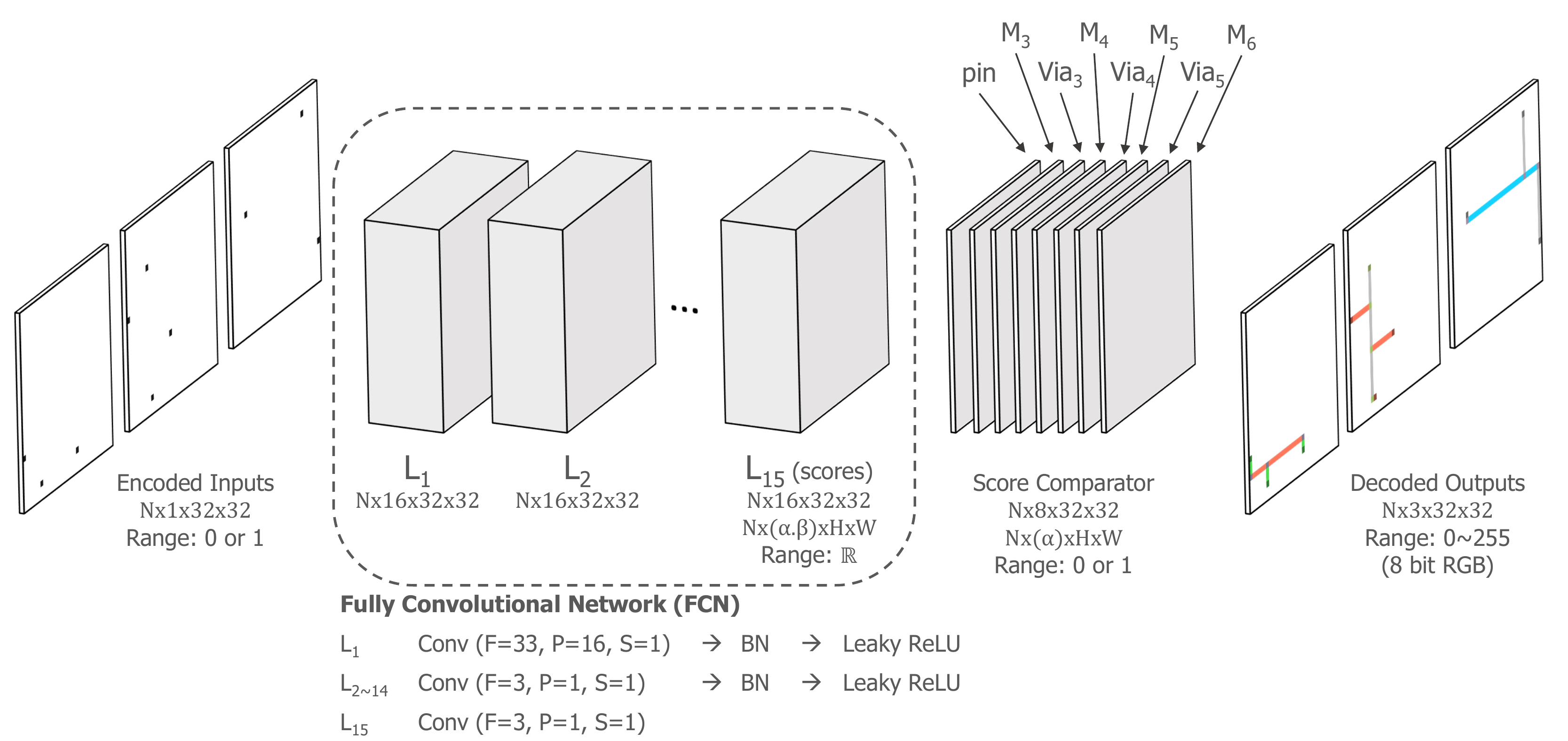}
  \caption{Model overview. The encoded layouts (with pins only) are fed to a 15 stage FCN which performs convolutions while preserving spatial dimensions. The filter size (F), padding (P) and stride (S) used in each stage is indicated. Scores at the end of the FCN are processed by a score comparator to generate the routed layouts (with pins, routes, and vias) which are visualized after decoding to 8-bit RGB.}
  \label{fig:network}
  \vspace{-0.15in}
\end{figure*}

\vspace{-0.05in}
\section{Model} \label{sec:model}
\vspace{-0.05in}
\noindent \textbf{Overview.} Our proposed model consists of a single, end-to-end, deep network using convolutions, which takes pin locations for a net as inputs and generates routes in one of the three wire class combinations, using layout design rules it learns from training data. We first present a binary scheme of encoding the input data in Section~\ref{sec:colorencoding}. The FCN model (see Figure~\ref{fig:network}) and its architectural details are covered in Section~\ref{sec:arch}. Then in Sections~\ref{sec:lossfn} and \ref{sec:training} we describe the loss function and the details of training respectively.

\vspace{-0.05in}
\subsection{Binary Encoding Scheme} \label{sec:colorencoding}
\vspace{-0.05in}
\noindent In typical image generation problems such as generative adversarial nets (GAN) \cite{GAN2014}, at the point of weight initialization, networks are intrinsically capable to generate an arbitrary color at any pixel coordinate. Subsequently through the course of training, meaningful color/coordinate combinations must be learned. The standard 8-bit RGB color gamut has $2^{24}$ color representations (3 color channels, 8 bits each) which is significantly larger than what we need. Since our data only deals with 8 layout layers (4 route, 3 via, 1 pin) we can immediately set an upper bound on the representation combinations per pixel to $2^8$. To take advantage of this insight, we choose to encode our data on a binary basis per-pixel per-layer. Hence instead of representing images as tensors of shape $3 \times H \times W$ with range: $\{0-255\}$ (8-bit RGB), we encode our dataset as tensors of shape $8 \times H \times W$ with a binary range: $\{0, 1\}$. By doing so, we are able to formulate this as a layer-wise binary segmentation task, with cross entropy loss on the scores ($16 \times H \times W$) to let the network make a binary decision of the presence of each of the 8 layers at every pixel in the context window $H \times W$.

\vspace{-0.05in}
\subsection{Network Architecture} \label{sec:arch}
\vspace{-0.05in}
\noindent Figure~\ref{fig:network} illustrates our model with the activation volumes after each stage. The FCN has a total of 15 convolutional\footnote{Fully connected layers or excessively deep layers were avoided.} stages (see Section~\ref{sec:experiments} for further explanation on the choice of network depth). All convolutions except the last one are followed by batch normalization (BN) \cite{Ioffe2015} and leaky rectified linear unit (leaky ReLU) \cite{Maas2013}. The last convolution stage ($L_{15}$) outputs the scores, hence is not followed by BN and leaky ReLU as they would not affect the relative scores. Encoded inputs to the FCN are tensors of shape $N \times 1 \times H \times W$, where $N$ is the mini-batch size and $H=W=32$ (spatial dimensions). The first stage ($L_1$) uses $33\times33$ convolution with 16 filters, whereas stages $L_2$ through $L_{15}$ use $3\times3$ convolution with 16 filters. We believe the large receptive field at the head of the network allows a fast grasp of the overall pin locations and helps the network learn spatial information better, as will be explored in Section~\ref{sec:experiments}. Strides and padding for convolutions are such as to preserve the spatial dimension of the feature maps at each stage. Thus the activations after each stage are of shape $N \times 16 \times 32 \times 32$, or more generally $N \times (\alpha.\beta) \times H \times W$ where $\alpha$ is the number of layout layers to be learned, and $\beta$ is the number of segmentation classes (two in our case). Activations from the last FCN stage (scores) are passed through the score comparator to pick the class with a higher score. Thus scores of shape $N \times (\alpha.\beta) \times H \times W$ are reduced to layout maps of shape $N \times \alpha \times H \times W$ with each pixel being either `1' or `0' indicating the presence or absence of a layer at that spatial location. For visualization of the routes, we decode the score comparator outputs to 8-bit RGB.

\vspace{-0.05in}
\subsection{Loss Function} \label{sec:lossfn}
\vspace{-0.05in}
\noindent To implement the network objective as a layer-wise binary segmentation task wherein every pixel in every layer is classified into one of the two classes ($y=0$ or $y=1$), we perform simple but critical reshape operations as follows. At train time, scores from the FCN (tensor) of shape $N \times (\alpha.\beta) \times H \times W$ are first reshaped to a matrix of shape $N.H.W.\alpha \times \beta$. The corresponding labels (tensor) of shape $N \times \alpha \times H \times W$ are reshaped to a vector of length $N.H.W.\alpha$. We then use averaged cross entropy loss (combination of negative log likelihood and softmax) over the predicted score matrix and label vector, to train our model (see Eq.~\ref{eq:loss}).

\vspace{-0.05in}
\begin{equation}
\begin{aligned}
L = \frac{1}{N.H.W.\alpha}\sum_{N.H.W.\alpha}{k_y \left \{-\log{\frac{e^{S_{y}}}{\sum_{\beta}{e^{S_{\beta}}}}} \right \} } \\[2mm]
+ \lambda \sum_n{\sum_{\alpha\times\beta}\sum{w^2_{n_{\alpha\times\beta}}}}
\label{eq:loss}
\end{aligned}
\end{equation}
\vspace{-0.05in}


\noindent \textbf{Class imbalance.} Since a majority of pixels in the input labels are background ($y=0$) with very few active pixels ($y=1$), we observe that the network quickly learns to classify all pixels as background and struggles to learn further. To mitigate such sparse learning difficulties due to class imbalance, we use weighted cross entropy loss with weights $k{_{y = 1}=3}$ and $k{_{y = 0}=1}$.

\noindent \textbf{Regularization.} An $L_2$ regularization term was added to the loss to improve generalization. We sum over squared weights of 16 convolution filters ($\sum_{\alpha\times\beta}$) from each stage ($\sum_{n}$) and use a regularization coefficient $\lambda = 1\times10^{-5}$.

\vspace{-0.05in}
\subsection{Training and optimization} \label{sec:training}
\vspace{-0.05in}
\noindent We train the model from scratch, starting with default weight initialization. We use Adam \cite{kingma2014adam} with $\beta_1=0.9$, $\beta_2=0.999$, and $\epsilon=1\times10^{-8}$ to train the weights of the FCN components. We experiment with two different mini-batch sizes, viz. $N=10$ and $N=100$ with learning rates of $5\times10^{-5}$ and $5\times10^{-4}$ respectively (see Section~\ref{sec:experiments}). A mini-batch of 10 runs in approximately 90ms ($\sim$7.5 min per epoch) and a mini-batch of 100 takes about 600ms ($\sim$5 min per epoch) on a Tesla K80 GPU. We use PyTorch \cite{PyTorch} to train and implement the network.

\vspace{-0.05in}
\section{Experiments} \label{sec:experiments}
\vspace{-0.05in}

\begin{figure*}[t]
  \vspace{-0.05in}
  \centering
  \includegraphics[width=0.48\textwidth]{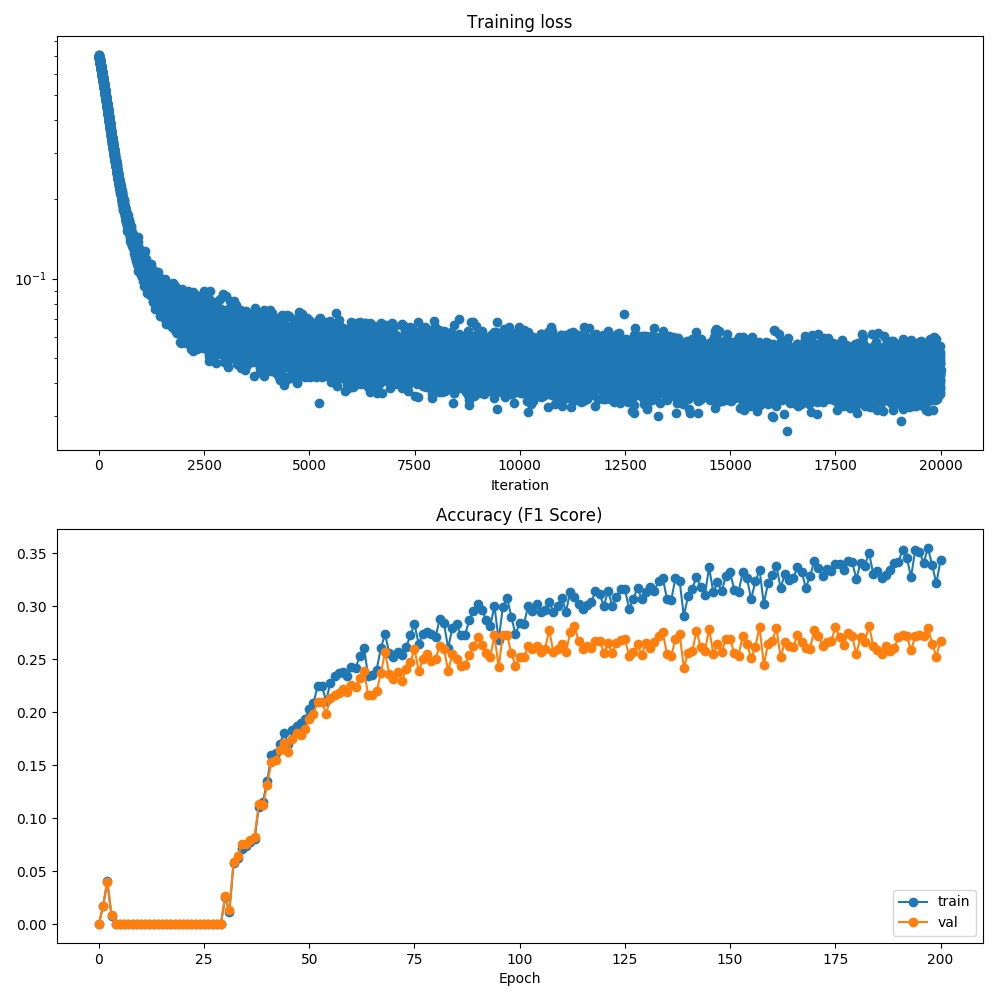}\quad
  \includegraphics[width=0.48\textwidth]{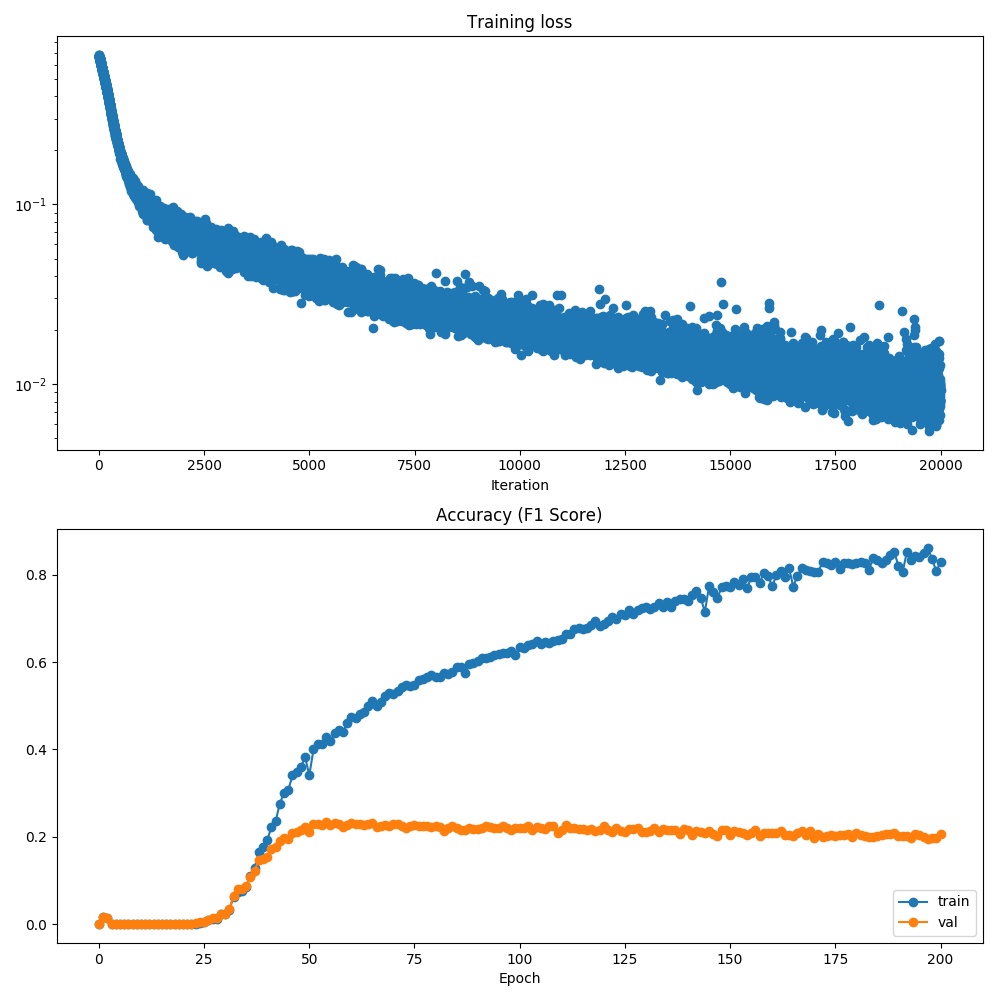}
  \caption{First stage experiment. We compare training curves with the first FCN stage using filters of size (i) F=3 (left) and (ii) F=33 (right). Small receptive field at the head of the network prevents the model from learning, as seen from the low train accuracy ($F_1\approx35\%$) on the left compared to the one on the right ($F_1\approx85\%$) after 200 epochs.}
  \label{fig:first_layer_experiment}
  \vspace{-0.15in}
\end{figure*}

\noindent \textbf{$\boldsymbol{F_1}$ score as accuracy metric.} Given the nature of our training data which is massively unbalanced towards one class ($y=0$), a raw metric comparing pixels of predicted and actual (ground truth) layouts will easily present an unreasonably high pixel-wise accuracy ($\sim$98\% in our case) even if the model incorrectly predicts all pixels to be background. We instead use the $F_1$ score metric that uses an equally weighted harmonic mean of precision and recall based confusion matrix. Precision is a measure of true positives among all pixels predicted positive. Recall is a measure of true positives among all pixels ground truth positive.
\vspace{-0.15in}
{\small \begin{verbatim}
### Pseudocode
# tp = true positives, tn = true negatives
# fp = false positives, fn = false negatives  
precision = tp / (tp+fp)
recall = tp / (tp+fn)
accuracy = (tp+tn) / (tp+tn+fp+fn)
f1_score = 2 * (precision*recall) / 
            (precision+recall)
\end{verbatim}}

\noindent \textbf{Network depth and receptive field.} We choose the FCN depth such as to have the overall receptive field of the network cover the entire input image ($32\times32$ pixels). Assuming all stages use $3\times3$ unit-strided convolutions, $3 + 2 \times (n_{stages}-1) > 32$, we would need at least $n_{stages} = 15$ to allow the network to learn reasonably well. We observe that reducing the FCN depth below 15 makes it difficult for the model to perfectly overfit ($F_1$ = 100\%) even a tiny dataset of 4 train samples. Note that our model uses $33\times33$ convolutions only in the first stage (see Figure~\ref{fig:network}).



\vspace{-0.05in}
\noindent \textbf{Significance of $\boldsymbol{F=33}$ in the first FCN stage.} As discussed in Section~\ref{sec:arch}, we use $33\times33$ convolutional filters in the first FCN stage as we believe this large receptive field at the head of the network helps the model quickly learn the correspondence between spatial spread of pins in the input and crucial route decisions such as wire class combinations, track usage, and branch-leg assignment. To demonstrate the significance of this architecture, we compare two models differing only in their first FCN stages, viz. (i) $F=3$, and (ii) $F=33$. Rest of the model is unchanged. For this experiment we train the two models on a subset of our dataset (1000 train and 200 validation samples), using mini-batches of 10, learning rate of $5\times10^{-5}$ and regularization strength of $1\times10^{-5}$. From Figure~\ref{fig:first_layer_experiment} we see that learning stagnates around $F_1=35\%$ after 200 epochs with $F=3$, whereas the model is able to quickly overfit the small dataset fairly well ($F_1=85\%$) with $F=33$. Interestingly, the increasing gap between training and validation accuracies is due to the small dataset size used for this experiment, causing the model to overfit to it.

\begin{figure*}[t]
\vspace{-0.05in}
\begin{center}
  \includegraphics[width=0.48\textwidth]{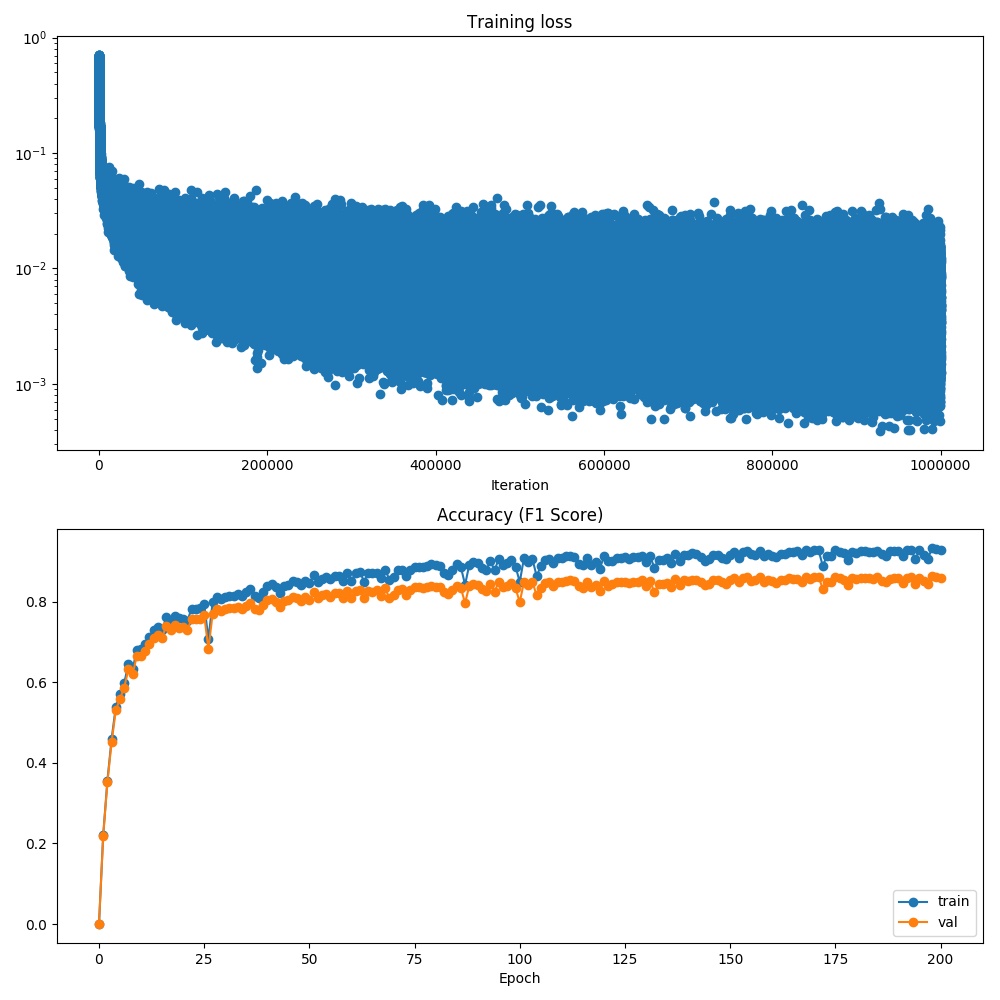}\quad
  \includegraphics[width=0.48\textwidth]{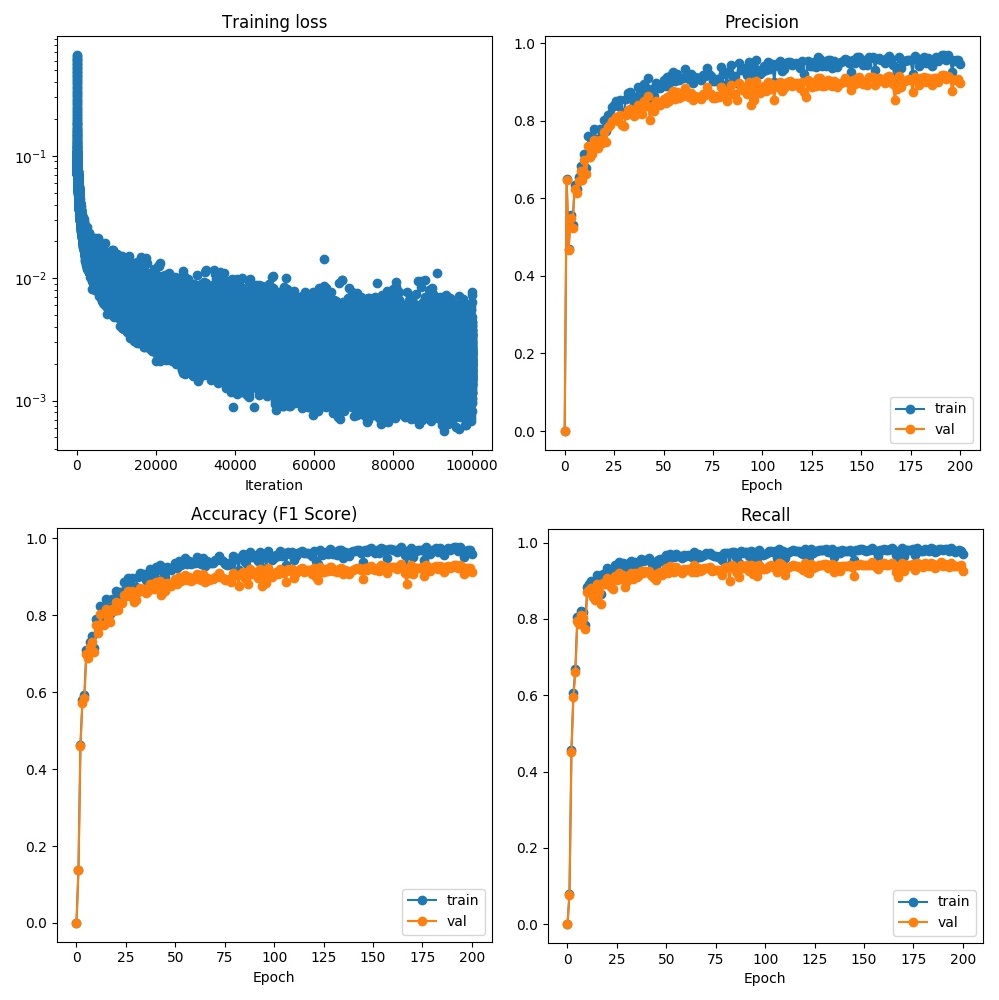}
  \caption{Loss and accuracy curves on training and validation sets from final training. Left shows the case with mini-batch of 10, learning rate of $5\times10^{-5}$. Right shows the case with mini-batch of 100, learning rate of $5\times10^{-4}$ (linear scaling). Precision and recall curves are included for the second case. Best accuracy was recorded at $F_1\approx97\%$ on train set and $F_1\approx92\%$ on validation set within 200 epochs.}
  \label{fig:final_training}
\end{center}
\vspace{-0.25in}
\end{figure*}

\noindent \textbf{Final training with different mini-batches.} Taking inspiration from \cite{priya17}, we conduct the final training using two different mini-batch sizes, viz. $N=10$ and $N=100$. The linear scaling rule \cite{priya17} suggests adjusting the learning rate linearly as a function of mini-batch size. This makes intuitive sense because we make fewer iterative steps per epoch with a larger mini-batch, hence the step size (learning rate) needs to be proportionally larger. We set the learning rates to $5\times10^{-5}$ and $5\times10^{-4}$ for mini-batches of 10 and 100 respectively, which worked reasonably well in our hyperparameter tuning experiments. To overcome overfitting seen in Figure~\ref{fig:first_layer_experiment}, we now use our complete dataset of 50,000 train and 10,000 validation samples for training. Figure~\ref{fig:final_training} shows the loss and accuracy curves (on both train and validation sets) for the two mini-batches. The overall trends look comparable. With mini-batch of 10 (left plot), the model achieves accuracies $F_1\approx90\%$ and $F_1\approx82\%$ on the train and validation sets after 200 epochs, and takes about 45 epochs to reach validation accuracy of $80\%$. In contrast, with mini-batch of 100 (right plot) the model achieves $F_1\approx97\%$ and $F_1\approx92\%$ on the train and validation sets after 200 epochs, and takes only 20 epochs to reach validation accuracy of $80\%$. Total train time significantly improved from 25 hours ($\sim$7.5 min per epoch) with mini-batch of 10 to 16 hours ($\sim$5 min per epoch) with mini-batch of 100, on a Tesla K80 GPU. The curves show good generalization of the model on the validation set with only a small gap ($<5\%$) between training and validation accuracies. Also included are the precision and recall curves for the second case.

\vspace{-0.05in}
\section{Results} \label{sec:results}
\vspace{-0.05in}

\begin{figure}[ht]
\vspace{-0.05in}
  \centering
  \includegraphics[width=0.32\linewidth]{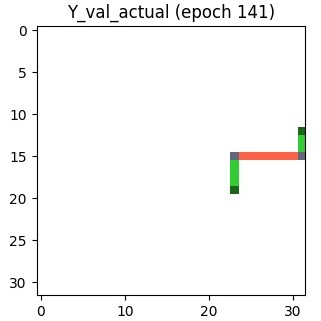}
  \includegraphics[width=0.32\linewidth]{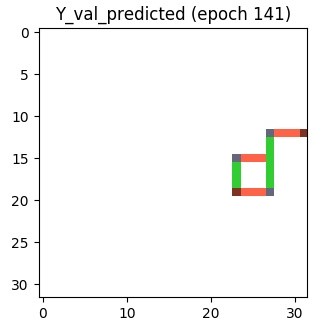}
  \includegraphics[width=0.32\linewidth]{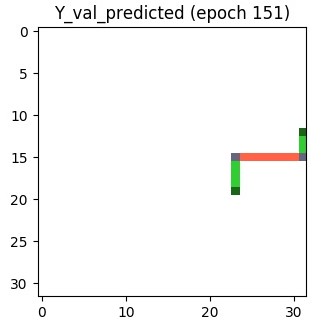}
  \caption{Learning example from the validation set. Left image shows actual layout (ground truth), center and right images show model predicted layouts after 141 and 151 epochs respectively.}
  \label{fig:learning_example}
\vspace{-0.05in}
\end{figure}

\noindent To gain insight into the learning process, we show a routed net example from the validation set with two pins (see Figure~\ref{fig:learning_example}). The left image shows the actual routed layout (ground truth), the center and right images show the predicted model outputs after 141 and 151 epochs of training respectively. We first notice that the model grasps orthogonality of adjacent metal layers, assigning $M_3$ (green) to vertical tracks and $M_4$ (red) to horizontal tracks only. Second, the model learns to connect different wire classes using vias at intersections. Third, the model learns to assign a branch to the dominant direction and legs to the non-dominant direction, however since the pins are roughly evenly spaced in either direction, the model attempts a vertical branch and horizontal legs. Eventually (after 10 epochs) it learns a more optimal way by using a horizontal branch and vertical legs to route, thus matching our ground truth expectation.

\begin{figure*}[t]
\vspace{-0.15in}
  \centering \quad
  \includegraphics[width=0.45\textwidth]{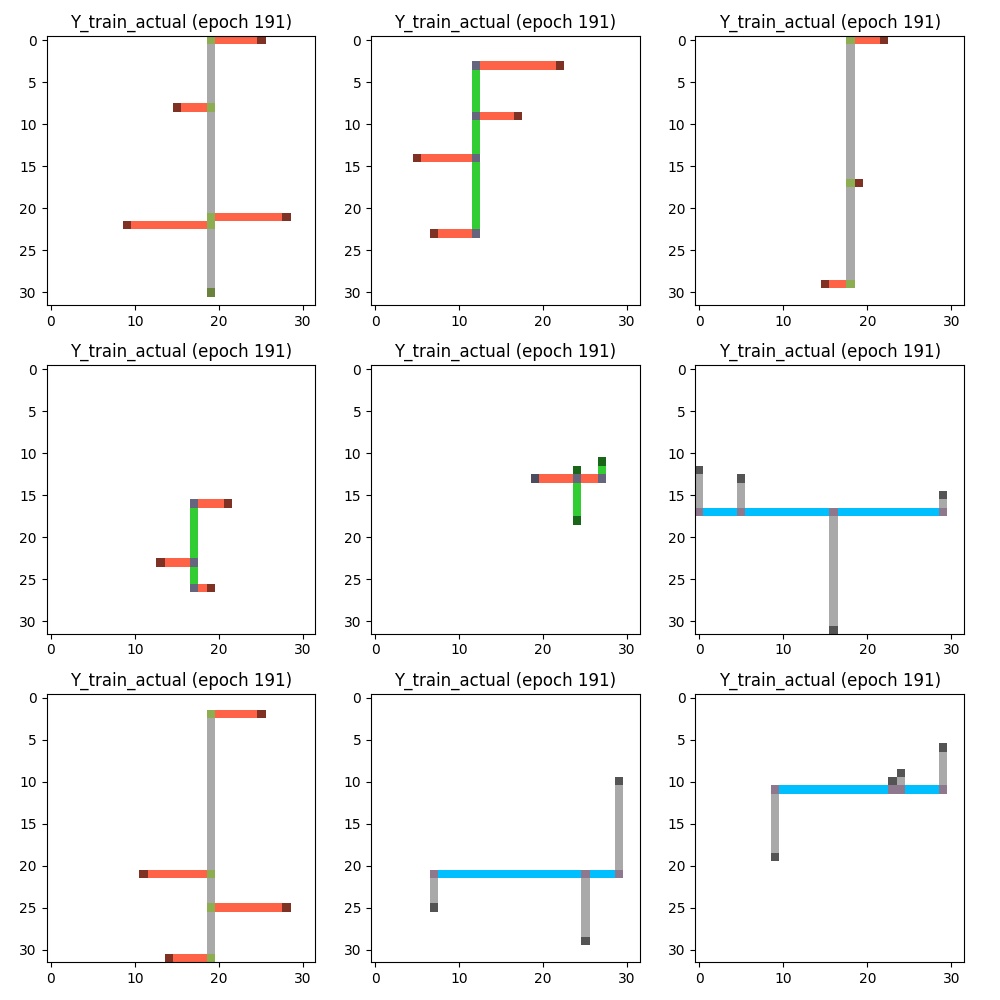}\quad
  \includegraphics[width=0.45\textwidth]{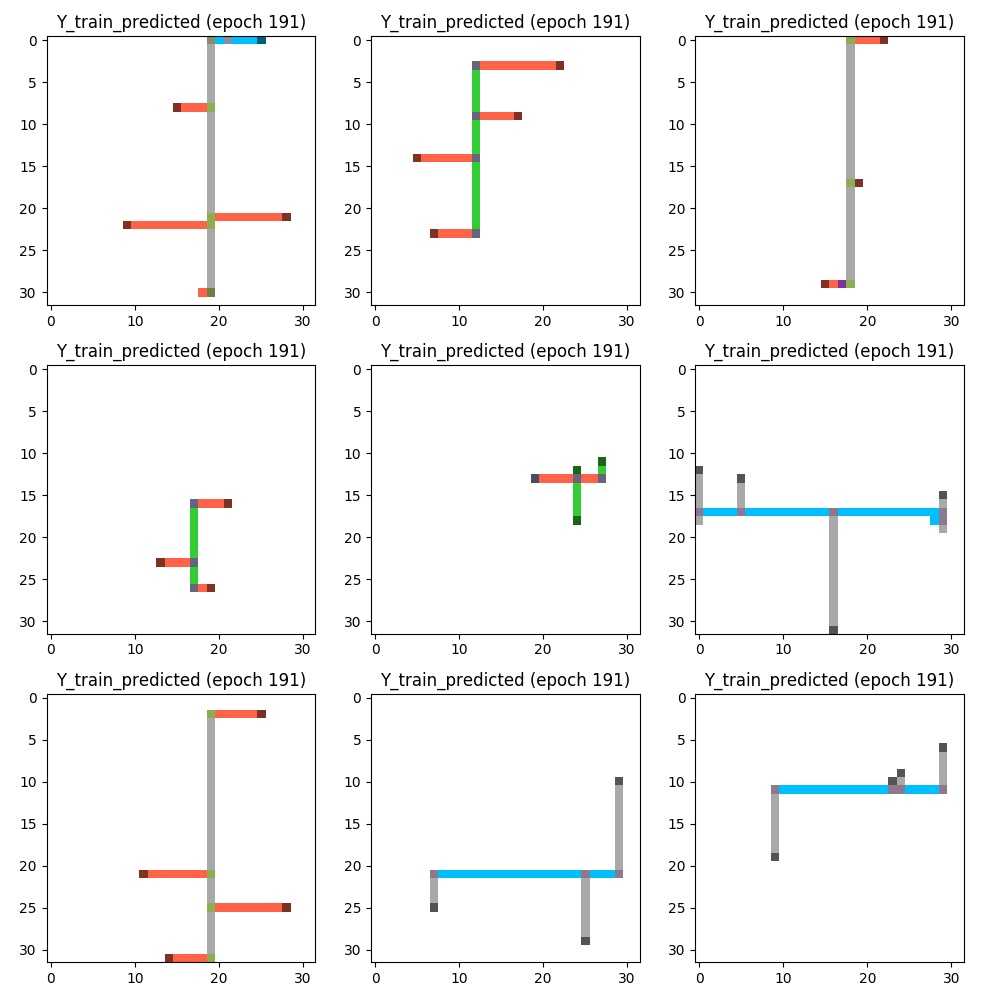}\newline\newline
  \includegraphics[width=0.45\textwidth]{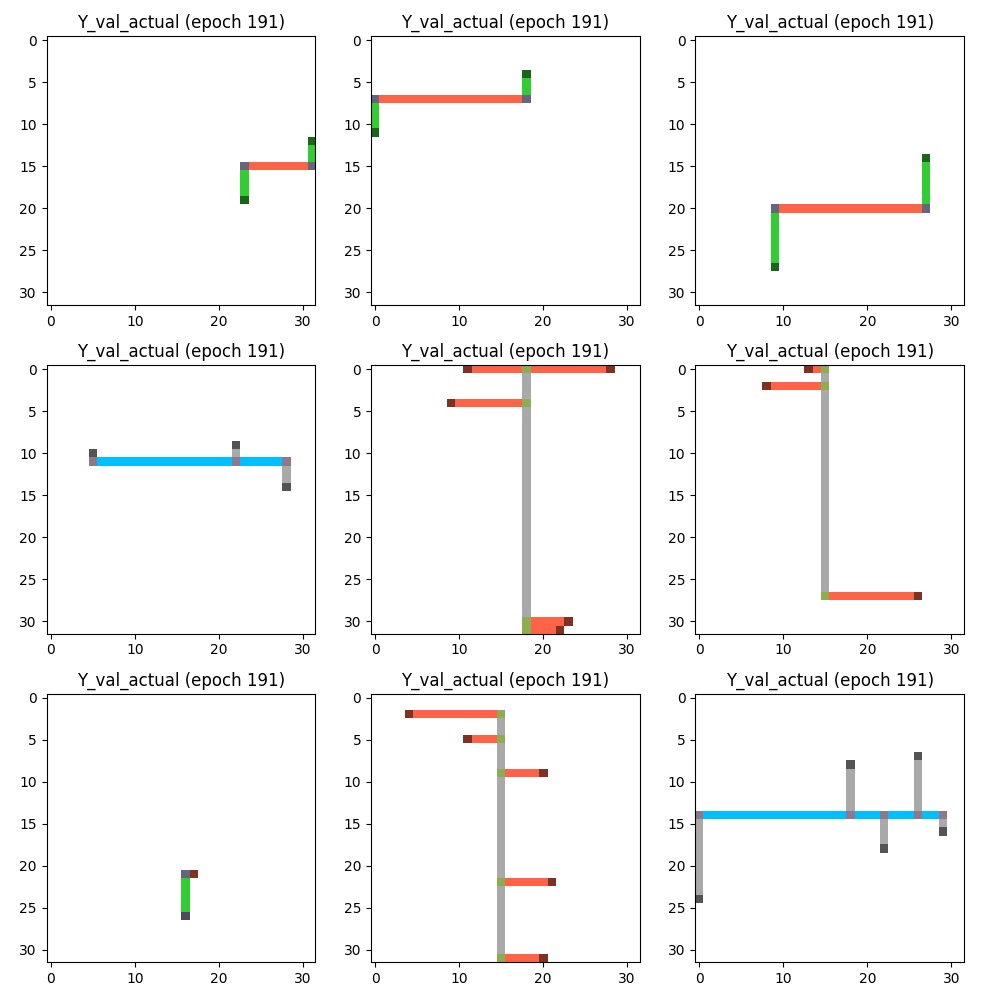}\quad
  \includegraphics[width=0.45\textwidth]{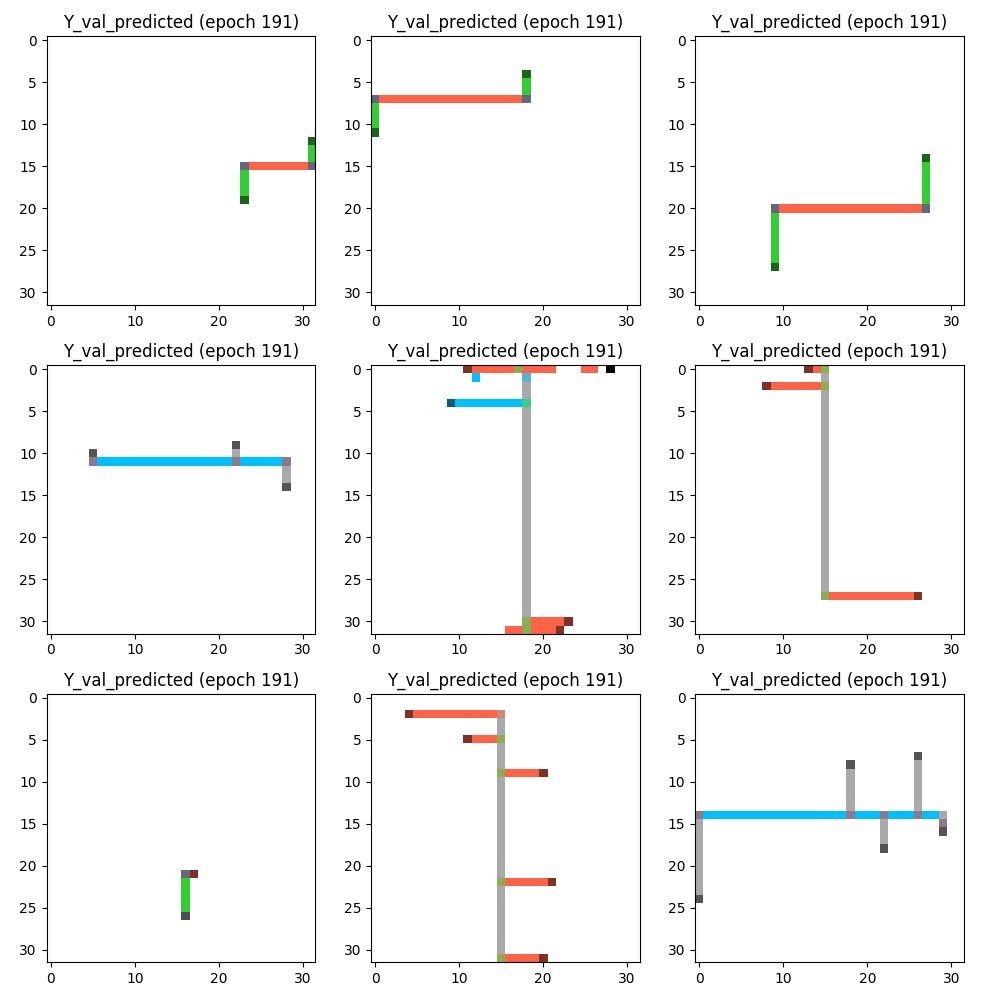}
  \caption{Examples from training (top row) and validation (bottom row) datasets showing the actual (ground truth) layouts on the left and the corresponding predicted (model routed) layouts on the right, after 191 epochs of training. The FCN model demonstrates good ability to learn layout design rules intrinsic to the dataset. Wire class color coding: $M_3$=green, $M_4$=red, $M_5$=grey, $M_6$=blue.}
  \label{fig:train_comparison}
\vspace{-0.15in}
\end{figure*}

Figure~\ref{fig:train_comparison} presents some randomly sampled examples from the training and validation sets, routed by our model after 191 epochs of training. As seen from the actual (ground truth) and predicted layouts, the network does well in (i) learning the identity mapping for pins, (ii) assigning vias to connect metals in adjacent layers, (iii) identifying the correct wire class combination from the overall pin configuration, and (iv) choosing the optimal track positions for branch and legs. It uses lower metals to route pins that are closer, and higher metals for widespread pins. In some cases, however, the routing is not perfect and the model misses connections or adds routes at undesired locations. We typically notice a higher error rate when layouts have more pins. This could likely be improved if we increase the ratio of training samples in the dataset containing more pins. While there is room for further improvement, the FCN model shows good overall ability to learn the layout design rules intrinsic to the dataset used for training.

\noindent \textbf{Future work.} Due to the complexity involved in routing real IC layouts, there are several requirements that need to be addressed. A few worth mentioning are (1) routing multiple nets in the presence of previously occupied tracks, (2) using a finer grid of valid metal tracks customized per wire class, (3) training on more complex route configurations such as trunk-branch-leg, (4) adding driver / receiver awareness to pins, (5) adding dedicated pin layers for each wire class, (6) supporting routing of a bigger layout in segments using a sliding context window, (7) integrating timing models for timing-driven routing, (8) training on industry standard layouts after converting to our layer-encoded binary standard. Some of these requirements may be implemented by direct scaling of the model and/or training with dense datasets containing more design rules of interest. However, it is plausible that other requirements may warrant more sophisticated architectures, possibly combining several neural models to construct a complex neural system to route ICs.

\vspace{-0.10in}
\section{Conclusion}
\vspace{-0.05in}
\noindent Inspired by the challenges facing circuit layout routing and optimization, and the recent advances in the field of convolutional neural networks, we introduced a unique approach to routing ICs using deep, fully convolutional networks. To explore learnability of layout design rules by our FCN model, we created our own dataset based on pre-defined layout constraints. We then implemented an encoding scheme to efficiently represent inputs to the model. The proposed FCN architecture efficiently learns to route a single net under set design constraints. Our model achieves good performance with training accuracy of $F_1\approx97\%$ and validation accuracy of $F_1\approx92\%$ within 200 epochs.

\vspace{-0.05in}
\section{Acknowledgements}
\vspace{-0.05in}
\noindent We thank Nishith Khandwala and Wenbin Huang for helpful comments and discussion. We gratefully acknowledge CS231N staff (Stanford) and Google Cloud for the educational GPU credits used towards this work.

\pagebreak

{\small
\bibliographystyle{ieee}
\bibliography{egbib}
}

\end{document}